\documentclass[10pt,twocolumn,letterpaper]{article}

\usepackage{cvpr}
\usepackage{times}
\usepackage{epsfig}
\usepackage{graphicx}
\usepackage{amsmath}
\usepackage{amssymb}

\usepackage{symbols}

\usepackage{color}
\usepackage{multirow}
\usepackage{url}
\usepackage{epstopdf}
\usepackage{ctable}
\usepackage{multirow}
\usepackage{array}
\usepackage{appendix}
\usepackage{subfigure}
\usepackage{caption}

\usepackage[pagebackref=true,breaklinks=true,letterpaper=true,colorlinks,bookmarks=false]{hyperref}

\cvprfinalcopy 

\def\cvprPaperID{3325} 

\newcommand{\specialcell}[2][c]{%
	  \begin{tabular}[#1]{@{}l@{}}#2\end{tabular}}

\ifcvprfinal\fi
\begin{document}

\title{Convolutional Image Captioning}
\author{Jyoti Aneja\thanks{~Denotes equal contribution.}, Aditya Deshpande\footnotemark[1], Alexander Schwing\\
University of Illinois at Urbana-Champaign\\
{\tt\small \{janeja2, ardeshp2, aschwing\}@illinois.edu}
}
\maketitle

\begin{abstract}
Image captioning is an important but challenging task, applicable to virtual assistants, editing tools, image indexing, and support of the disabled. Its challenges are due to the variability and ambiguity of possible image descriptions. In recent years significant progress has been made in image captioning, using Recurrent Neural Networks powered by long-short-term-memory (LSTM) units. Despite mitigating the vanishing gradient problem, and despite their compelling ability to memorize dependencies, LSTM units are complex and inherently sequential across time. To address this issue, recent work has shown benefits of convolutional networks for machine translation and conditional image generation~\cite{S2S,Oord2016,VaswaniAttn}. Inspired by their success, in this paper, we develop a convolutional image captioning technique. We demonstrate its efficacy on the challenging MSCOCO dataset and demonstrate performance on par with the baseline, while having a faster training time per number of parameters. We also perform a detailed analysis, providing compelling reasons in favor of convolutional language generation approaches. 
\end{abstract}

\section{Introduction}

Image captioning, \ie, describing the content observed in an image, has
received a significant amount of attention in recent years. It is 
applicable in various scenarios, \eg, recommendation in editing applications,
usage in virtual assistants, for image indexing, and support of the disabled. 
It is also a basic ingredient for more complex operations such as 
storytelling~\cite{HaoStoryTelling} and visual summarization~\cite{VenugopalanVidSummary}.

\begin{figure}[t]
 \vspace{-0.5cm}
 \begin{center}
 \includegraphics[width=0.95\linewidth]{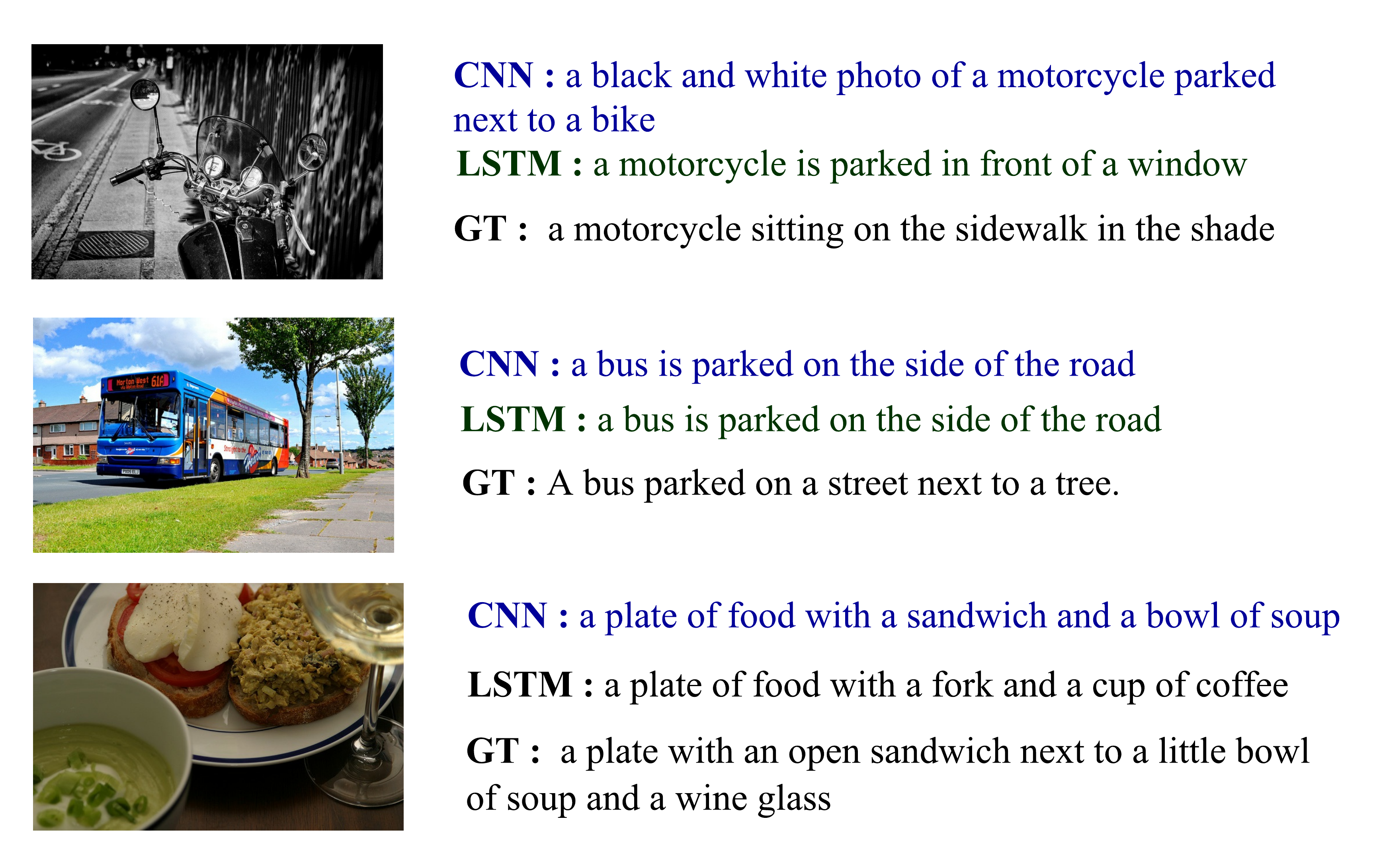}
 \end{center}
  \vspace{-0.8cm}
 \caption{Examples of captions generated by a standard LSTM-based network, our convolutional network (CNN) and ground-truth (GT) captions by a human annotator.}
 \label{fig:teaser}
  \vspace{-0.55cm}
 \end{figure}

In recent years, with the availability of larger
datasets, deep neural network (DNN) based methods have been shown to achieve impressive
results on image captioning tasks~\cite{neuraltalk, show_tell}. These techniques are 
largely based on recurrent neural nets (RNNs), often powered by a 
Long-Short-Term-Memory (LSTM)~\cite{HochreiterLSTM} component which mitigates 
the issue of vanishing or exploding gradients~\cite{PascanuDiffRNN}. 
 \figref{fig:teaser} shows examples of captions generated by an 
LSTM-based network.

LSTM nets have been considered as the de-facto standard for 
vision-language tasks of image captioning~\cite{XinleiCaption,
neuraltalk, show_tell, show_attend_tell}, visual question answering~\cite{VQA,ShihVQA}, 
question generation~\cite{JainQues,MostafazadehMDM16}, situation
recognition~\cite{Mallya17}, and visual dialog~\cite{DasVisDiag}, due to their compelling ability to memorize long-term dependencies through a memory cell. 
However, the complex addressing and overwriting mechanism combined with inherently sequential processing, and significant storage required due to back-propagation through time 
(BPTT), poses challenges during training. Also, in contrast to CNNs, that are non-sequential, LSTMs often require more careful engineering, when considering a novel task. Previously, CNNs have not matched up to the LSTM 
performance on vision-language tasks. 

\begin{figure}[!t]
 \vspace{-0.5cm}
\centering
\includegraphics[width=\linewidth]{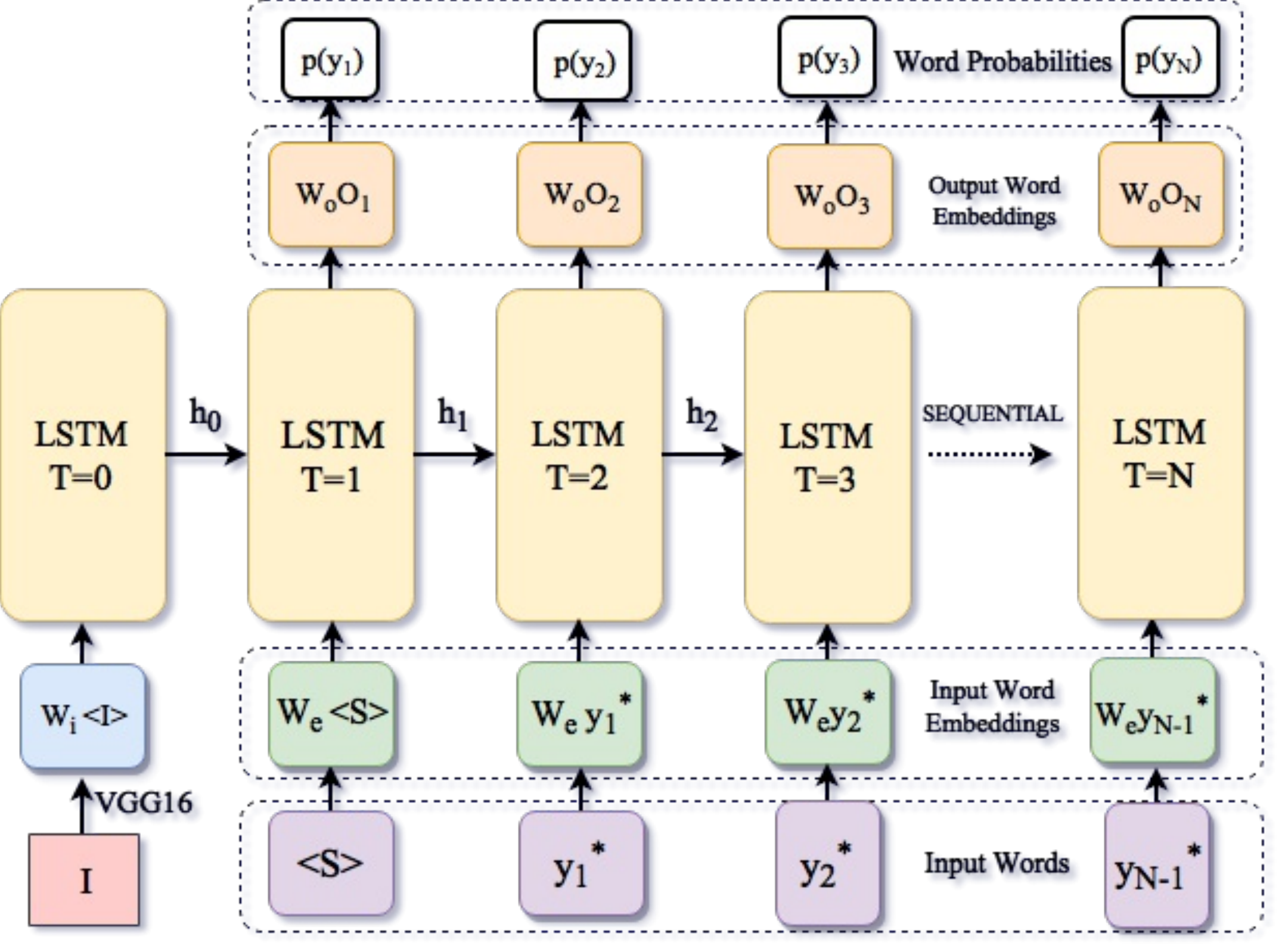}
\vspace{-0.8cm}
\caption{A sequential RNN powered by an LSTM cell. At each time step 
output is conditioned on the previously generated word, the image is
fed at the start only.}
\label{fig:HighLevel_lstm}
\vspace{-0.5cm}
\end{figure}

Inspired by the recent successes of convolutional architectures on 
other sequence-to-sequence tasks -- conditional image 
generation~\cite{Oord2016}, machine translation~\cite{S2S,VaswaniAttn} 
-- we study convolutional architectures for the  
task of image captioning. To the best of our knowledge, ours is the 
first convolutional network for image captioning that compares 
favorably to LSTM-based methods. On the MSCOCO dataset we match 
performance of the LSTM with a .952 CIDEr score and get a better 
BLEU-4 score of .316 (see  \tabref{tab:beam_search}). 

Our key contributions are: \textbf{a)} A convolutional (CNN-based) 
image captioning method that shows comparable performance to an LSTM based method on standard 
metrics (\secref{sec:results_metrics}, \tabref{tab:metrics} and \tabref{tab:beam_search});
\textbf{b)} Improved performance with a CNN model that uses attention 
mechanism to leverage spatial image features. With attention, we 
outperform the current attention baseline and qualitatively demonstrate that 
our method finds salient objects in the image. (\figref{fig:attention}, 
\tabref{tab:beam_search}); \textbf{c)} We analyze the characteristics of CNN and 
LSTM nets and provide useful insights such as -- CNNs produce more entropy 
(useful for diverse predictions), better classification accuracy, and do not 
suffer from vanishing gradients (\secref{sec:results} and  \figref{fig:analysis},~\ref{fig:words} 
and \ref{fig:grad}). We evaluate our architecture on the 
challenging MSCOCO~\cite{MSCOCO} dataset, and compare it to an LSTM~\cite{neuraltalk} 
and an LSTM+Attention baseline~\cite{show_attend_tell}.

The paper is organized as follows:  \secref{sec:notation} describes our 
notation,  \secref{sec:rnn} briefly reviews the RNN/LSTM based approach, 
 \secref{sec:fconv} describes our convolutional method and 
 \secref{sec:arch} gives the details of our CNN architecture,  
 \secref{sec:results} contains our results and 
 \secref{sec:relwork} discusses previous work on image captioning.


\section{Problem Setup and Notation}
\label{sec:notation}

For image captioning, we are given an input image $I$ 
and we are required to generate a sequence of words 
$y = (y_1, \ldots, y_N)$. The possible words $y_i\in\cY$ 
at time-step $i$ are subsumed in a discrete set $\cY$ of 
options.  The number of possible options $|\cY|$ easily 
reaches several thousands. This set of options $\cY$ often 
contains special tokens that denote a start token ($<$S$>$), 
an end of sentence token ($<$E$>$), and an unknown token ($<$UNK$>$)
which refers to all words not in $\cY$.

Given a training set $\cD = \{(I, y^\ast)\}$ which contains pairs $(I,y^\ast)$ 
of input image $I$ and corresponding ground-truth caption 
$y^\ast = (y^\ast_1, \ldots,y^\ast_N)$, consisting of words 
$y^\ast_i\in\cY$, $i\in\{1,\ldots, N\}$, 
we maximize \wrt parameters $w$, a probabilistic model $p_w(y_1, \ldots, y_N | I)$.  


A variety of probabilistic models have been considered (\secref{sec:relwork}), from 
hidden Markov models~\cite{Yang11} to recurrent neural networks. 
First, we briefly review the inference and learning of 
RNN-based approaches (\secref{sec:rnn}) and then delve into 
our convolutional approach (\secref{sec:fconv}).

\section{RNN Approach}
\label{sec:rnn}

An illustration of a classical RNN architecture for image 
captioning is provided in  \figref{fig:HighLevel_lstm}. It 
consists of three major components, all of which contain 
trainable parameters: the input word embeddings, the 
sequential LSTM units containing the memory cell, and the 
output word embeddings.

\noindent{\bf Inference.} RNNs sequentially predict one word 
at a time,  from $y_{1}$ up to $y_{N}$. At every time-step $i$, 
a conditional probability distribution $p_{i,w}(y_i|I)$, which 
depends on parameters $w$, is predicted (see top of  
\figref{fig:HighLevel_lstm}). During inference, we typically choose 
the word $y_i$ with the highest probability $p_{i,w}(y_i|I)$. 
The sentence terminates once the end of sentence token is chosen, 
or after a maximum of $N$ steps.

For modeling $p_{i,w}(y_i|I)$, in the spirit of auto-regressive 
models, the dependence of word $y_i$ on its ancestors $y_{<i}$ 
is implicitly captured by a hidden representation $h_i$ 
(see arrows in  \figref{fig:HighLevel_lstm}). Formally, the 
probability is computed via
\be
p_{i,w}(y_i|h_i,I) = g_w(y_i, h_i, I),
\ee
where $g_w$ can be any differentiable function/deep net. This 
function may depend directly on the image $I$. However, classical 
image captioning techniques usually encode the image into the 
hidden representation $h_0$ (\figref{fig:HighLevel_lstm}). 

Importantly, RNNs just like other auto-regressive models are described by a
recurrence relation which governs computation of the hidden state $h_i$ based
on its values at the previous time instant, $h_{i-1}$, as well as the word $y_{i-1}$, predicted at the
previous time step: 
\be 
h_{i} = f_w(h_{i-1}, y_{i-1}, I).
\label{eq:RecurrenceInf} 
\ee 
Again, $f_w$ can be any differentiable function. For image captioning, 
long-short-term-memory (LSTM)~\cite{HochreiterLSTM} nets and variants thereof based 
on gated recurrent units (GRU)~\cite{ChoGRU}, or forward-backward LSTM
nets are used here.

The recurrent function given in \equref{eq:RecurrenceInf} generally doesn't 
operate directly on one-hot input vectors $y_{i}$. Instead, input words are 
encoded into a vectorial representation via an embedding layer (input word 
embeddings in \figref{fig:HighLevel_lstm}). Note that the functions 
$g_w$ and $f_w$ are identical across time.

\noindent{\bf Learning.} 
Following classical supervised learning, it is common to find the parameters 
$w$ of the word embeddings and the LSTM unit by minimizing the negative log-likelihood of the training data $\cD$, 
\ie, we optimize: 
\be \min_w -\sum_{(I,y^\ast)\in\cD}\sum_{i=1}^N
 \ln p_{i,w}(y^\ast_i | h_i, I).  \label{eq:RNNObjective} \ee

To compute the gradient of the objective given in \equref{eq:RNNObjective},
back-propagation through time (BPTT) is the de-facto standard these days.
BPTT is necessary due to the recurrence relationship encoded in $f_w$ 
(\equref{eq:RecurrenceInf}), which is unrolled as illustrated in 
\figref{fig:HighLevel_lstm}. The gradients of the function $f_w$ at time 
$i$ depend on the gradients obtained in successive time-steps.

To avoid more complicated gradient flows through the recurrence relationship,
during training, it is common to use 
\be h_{i} = f_w(h_{i-1}, y^\ast_{i-1},I),  
\label{eq:RecurrenceTrain} 
\ee rather than the form
provided in \equref{eq:RecurrenceInf}. \Ie, during training, when computing the latent
representation $h_i$, we use the ground-truth symbol $y^\ast_{i-1}$ rather
than the prediction $y_{i-1}$. This is termed as teacher forcing. While this 
simplifies the gradient flow, it obviously causes the input statistics to differ
between train and test. This issue has been subject to recent work~\cite{Bengio15,goyal16}.

Although highly successful, RNN-based techniques suffer from some drawbacks.
First, the training process is inherently sequential for a particular
image-caption pair. This results from unrolling the recurrent relation in time. Hence, the output at time-step $i$ has a
true dependency on the output at $i-1$. Secondly, as we will show in our 
results for image captioning, RNNs tend to produce lower classification
accuracy (\figref{fig:analysis}), and, despite LSTM units, they still suffer 
to some degree from vanishing gradients (\figref{fig:grad}).

Next, we describe an alternative convolutional approach to image captioning 
which attempts to overcome some of these challenges.  

\begin{figure}[!t]
\vspace{-0.5cm}
\centering
\includegraphics[width=\linewidth]{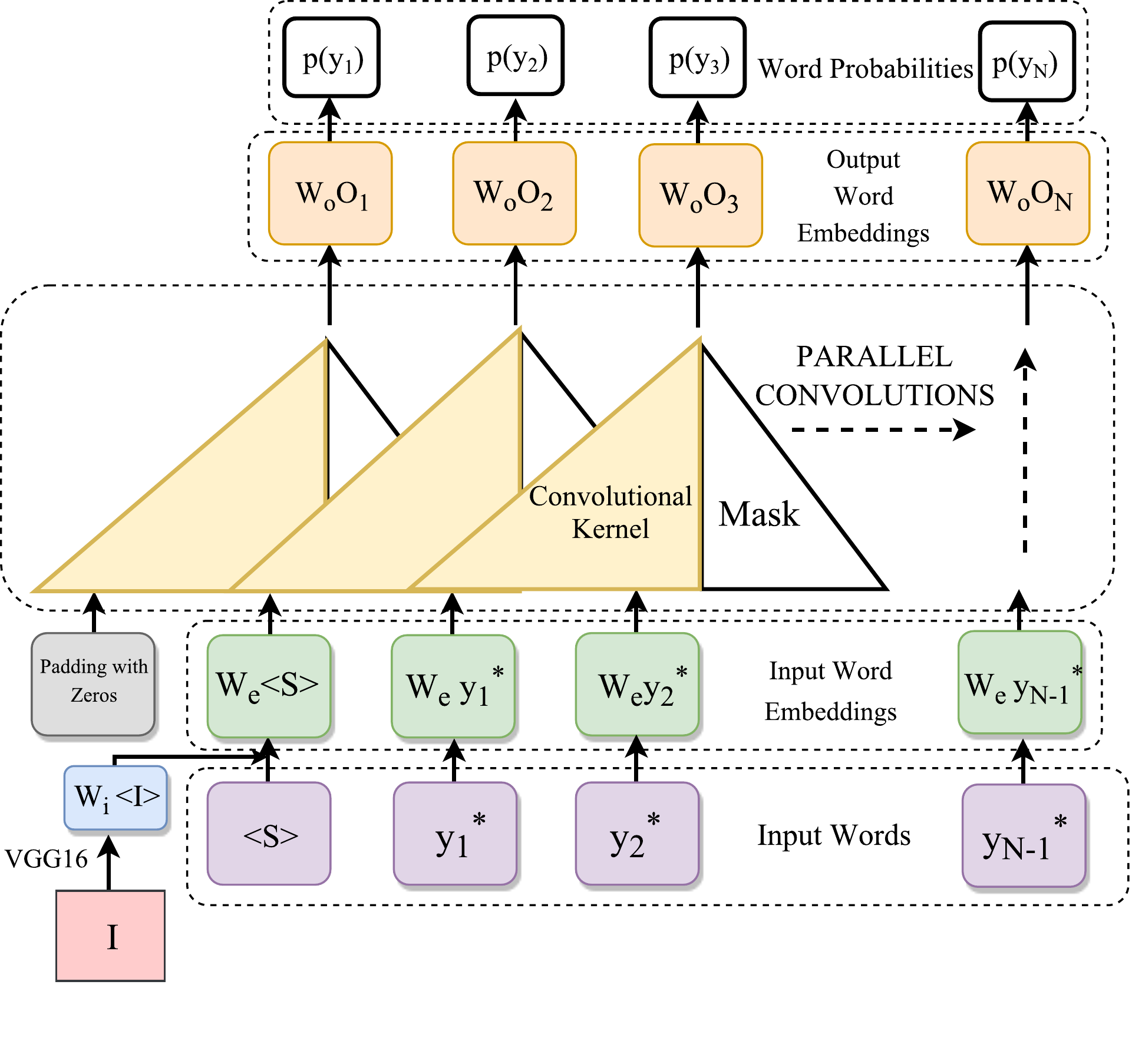}
\vspace{-1.3cm}
\caption{Our convolutional model for image captioning. We use a feed forward network with masked convolutions. Unlike RNNs, our model operates over all words in parallel.} 
\label{fig:HighLevel_fconv}
\vspace{-0.5cm}
\end{figure}


\section{Convolutional Approach}
\label{sec:fconv}

\begin{figure*}[!t]
\vspace{-0.5cm}
\centering
\includegraphics[width= 0.8\textwidth]{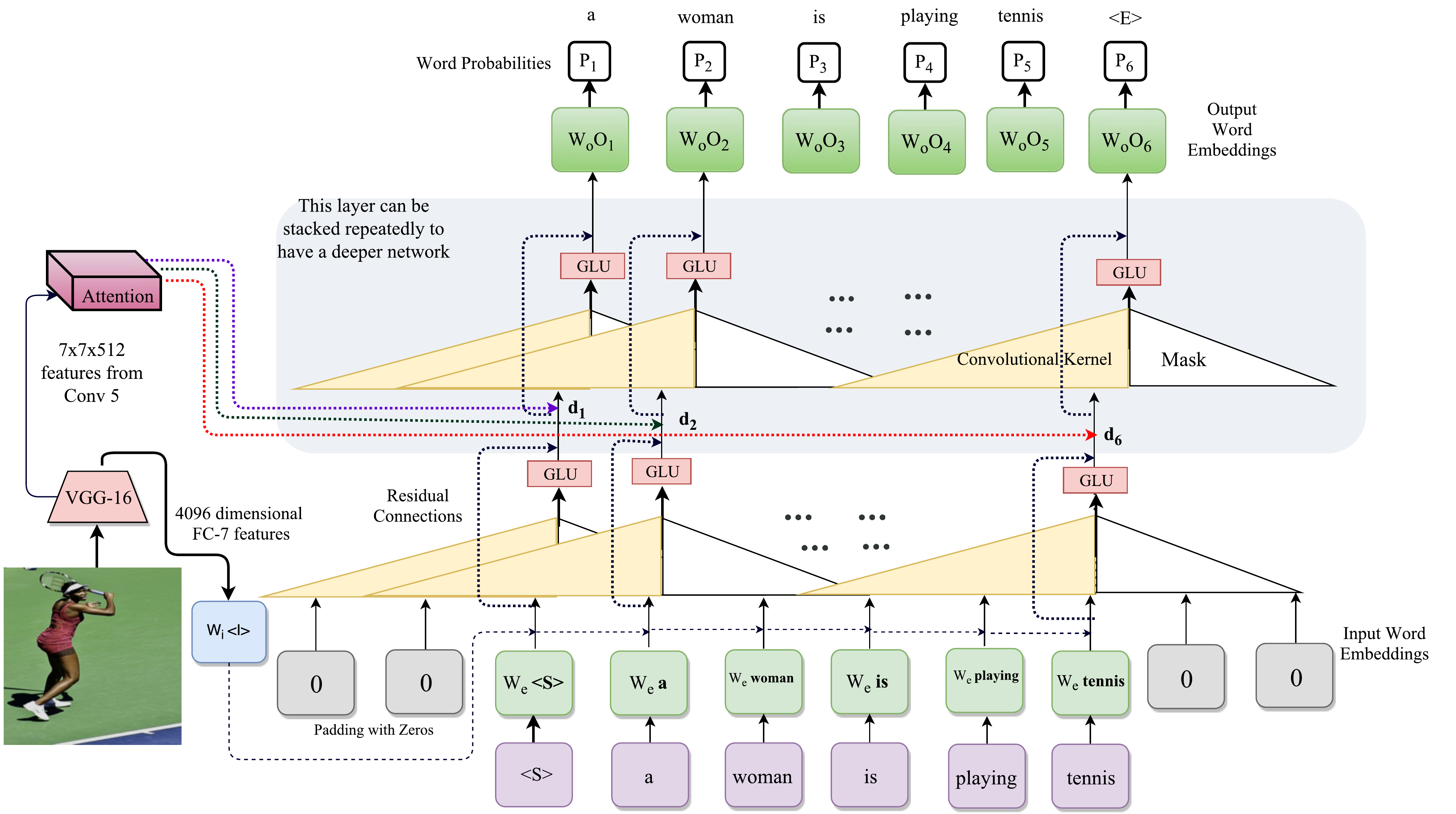}
\vspace{-0.3cm}
\caption{Our convolutional architecture for image captioning. Our architecture
has four components: (i) Input embedding layer, (ii) Image embedding, (iii) 
Convolutional module and (iii) Output embedding layer. Details of these components can be found in \secref{sec:arch}.}
\label{fig:Detailed}
\vspace{-0.5cm}
\end{figure*}

Our model is based on the convolutional machine translation model used 
in~\cite{S2S}. \figref{fig:HighLevel_fconv} provides an overview of 
our feed-forward convolutional (or CNN-based) approach for image 
captioning. As the figure illustrates, our technique contains three main 
components similar to the RNN technique. The first and the last components 
are word embeddings in both cases. However, while the center component 
contains LSTM or GRU units in the RNN case, masked convolutions are 
employed in our CNN-based approach. This component, unlike the RNN,
is feed-forward without any recurrent function. We briefly review 
inference and learning of our model. The problem setup and notation 
follows \secref{sec:notation}.

\noindent
{\bf Inference.} 
In contrast to the RNN formulation, where the probabilistic model is 
unrolled in time via the recurrence relation given in \equref{eq:RecurrenceTrain}, 
we use a simple feed-forward deep net, $f_w$, for modeling $p_{i,w}(y_i|I)$. 
Prediction of a word $y_{i}$  relies on past words $y_{<i}$ or their representations:
\be 
p_{i,w}(y_i| y_{<i}, I) = f_w(y_i, y_{<i}, I).
\label{eq:feedforward}
\ee

To disallow convolution operations from using information of future word 
tokens, we use masked convolutional layers that operate only on `past' 
data. This is similar to~\cite{S2S, Oord2016}. 

Inference can now be performed sequentially, one word at a time. Hence, 
inference begins with the start token $<$S$>$ and employs a 
feed-forward pass to generate $p_{1,w}(y_1|\emptyset,I)$. Afterwards,  
$y_1 \sim p_{1,w}(y_1|\emptyset,I)$ is sampled. Note that it is possible 
to retrieve the maximizing argument or to perform beam search. 
After sampling, $y_1$ is fed back into the feed-forward network 
to generate subsequent words $y_2$, \etc. Inference continues 
until the end token is predicted, or until we reach a fixed 
upper bound of $N$ steps. 

\noindent
{\bf Learning.} Similar to RNN training, we use ground-truth 
$y^\ast_{<i}$ for past words, instead of using the  
most recently generated ones. For prediction of word 
probability $p_{i,w}(y_{i}|y_{<i}^\ast,I)$, the considered feed-forward 
network  is $f_w(y_i,y^\ast_{<i}, I)$ and we optimize for 
parameters $w$ using a likelihood similar to 
\equref{eq:RNNObjective}.

Since there are no recurrent connections and all 
ground-truth words are available at any given time-step $i$, 
our CNN based model can be trained in parallel for all words. 
In Section \ref{sec:arch}, we provide details about 
the convolutional architecture used in our experiments.

\section{Architecture} 
\label{sec:arch}

The detailed architecture for our convolutional image captioning 
is illustrated in  \figref{fig:Detailed}. In this figure, we show 
a training iteration with input (ground-truth) words 
\{$y^\ast_{1}, \ldots, y^\ast_{5}$\} = $\{$ a, woman, is, playing, 
tennis $\}$. Additionally, we add the start token $<$S$>$
at the beginning and also, the end of sentence token $<$E$>$.

These words are processed as follows: (1) they pass through an 
input embedding layer; (2) they are combined with the image embedding; 
(3) they are processed by the CNN module; and (4)  the output
embedding (or classification) layer produces output probability 
distributions (see $\{p_{1}, \ldots, p_{6}\}$ at top of 
\figref{fig:Detailed}). Each of the four aforementioned steps 
is discussed below. 

\noindent
{\bf Input Embedding.} For consistency with RNN/LSTM baseline, 
we train (from scratch) an embedding layer  over one-hot encoded 
input words. We use $|\cY| = 9221$ and we embed the input words 
to $512$-dimensional vectors, same as the baseline. This embedding 
is concatenated to the image embedding (discussed next) and provided as 
input to the feed-forward CNN module.

\noindent
{\bf Image Embedding.} Image features for image $I$, are 
obtained from the fc7 layer of the VGG16 network~\cite{Simonyan14c}. 
The VGG16 is pre-trained on the  ImageNet dataset~\cite{Russakovsky2015}. 
We apply dropout, ReLU on the fc7 and use a linear layer to obtain a 
$512$-dimensional embedding. This is consistent with the image 
features used in the baseline LSTM method~\cite{neuraltalk}. 

\begin{table*}[!t]
\begin{center}
\setlength{\tabcolsep}{3pt}
\begin{tabular}{|l|cccccccc|cccccccc|}
\hline
\multirow{2}{*}{\bf{Method}} & \multicolumn{8}{c|}{\bf{MSCOCO Val Set}} & \multicolumn{8}{c|}{\bf{MSCOCO Test Set}} \\ 
& B1 & B2 & B3 & B4 & M & R & C & S 
& B1 & B2 & B3 & B4 & M & R & C & S \\ 
\hline
\hline
{\bf Baselines:}
& & & & & & & & 
& & & & & & & & \\ 
LSTM \cite{neuraltalk}
& {\bf .710} & .535 & .389 & .281 & {\bf .244} & {\bf .521} & {\bf .899} & .169
& {\bf .713} & {\bf .541} & {\bf .404} & {\bf .303} & {\bf .247} & {\bf .525} & {\bf .912} & .172 \\ 
LSTM + Attn (Soft) \cite{show_attend_tell}
& - & - & - & - & - & - & - & -
& .707 & .492 & .344 & .243 & .239 & - & - & - \\
LSTM + Attn (Hard) \cite{show_attend_tell}
& - & - & - & - & - & - & - & -
& .718 & .504 & .357 & .250 & .230 & - & - & - \\
\hline
\hline
{\bf Our CNN:}
& & & & & & & & 
& & & & & & & & \\
\begin{tabular}{ll} CNN \end{tabular}
& .693 & .518 & .374 & .268 &  .238 & .511 & .855 & .167 
& .695 & .521 & .380 & .276 & .241 & .514 & .881 & .171 \\ 
\begin{tabular}{ll} CNN & + Weight Norm.\end{tabular}
& .702 & .528 & .384 & .279 & .242 & .517 & .881 & .169 
& .699 & .525 & .382 & .276 & .241 & .516 & .878 & .170 \\
\begin{tabular}{ll} CNN & +WN +Dropout \end{tabular}
& .707 & .532 & .386 & .278 & .242 & .517 & .883 & .171 
& .704 & .532 & .389 & .283 & .243 & .520 & .904 & .173 \\
\begin{tabular}{ll} CNN & +WN +Dropout \\ & +Residual \end{tabular}
& .706 & .532 & .389 & {\bf.284} & {\bf .244} & .519 & {\bf .899} & {\bf.173} 
& .704 & .532 & .389 & .284 & .244 & .520 & .906 & {\bf.175} \\
\begin{tabular}{ll} CNN & +WN +Drop. \\ & +Res. +Attn \end{tabular}
& {\bf .710} & {\bf .537} & {\bf .391} & .281 & .241 & .519 & .890 & .171 
& .711 & .538 & .394 & .287 & .244 & .522 & {\bf .912} & {\bf.175} \\
\hline
\end{tabular}
\end{center}
\vspace{-0.5cm}
\caption{Comparison of different methods on standard evaluation metrics: BLEU-1 (B1), 
BLEU-2 (B2), BLEU-3 (B3), BLEU-4 (B4), METEOR (M), ROUGE (R), CIDEr (C) and SPICE (S). 
Our CNN with attention (attn) achieves comparable performance (equal CIDEr scores on 
MSCOCO test set) to \cite{neuraltalk} and outperforms LSTM+Attention baseline of 
\cite{show_attend_tell}. We experiment with various configurations of our CNN. We 
start with a CNN comprising masked convolutions and fully connected layers only.
Then, we add weight normalization, dropout, residual connections and attention 
incrementally and show that performance improves with every addition. Here, for 
our CNN and \cite{neuraltalk} we use the model that obtains the best CIDEr 
scores on val-set and report its scores for the test set. We evaluate CIDEr scores 
after every training epoch for val-set. Note, the val set is not used for any 
hyper-parameter tuning other than choosing the best model for test set. 
For \cite{show_attend_tell}, we report all the available metrics for soft/hard attention 
from their paper (missing numbers are marked by -).}
\label{tab:metrics}
\end{table*}

\begin{table*}[!t]
\begin{center}
\setlength{\tabcolsep}{3pt}
\resizebox{\textwidth}{!}{
\begin{tabular}{|l|cccccccc|cccccccc|cccccccc|}
\hline
\multirow{2}{*}{\bf{Method}} & \multicolumn{8}{c|}{\bf{Beam Size=2}} & \multicolumn{8}{c|}{\bf{Beam Size=3}} &  \multicolumn{8}{c|}{\bf{Beam Size=4}} \\ 
& B1 & B2 & B3 & B4 & M & R & C & S 
& B1 & B2 & B3 & B4 & M & R & C & S 
& B1 & B2 & B3 & B4 & M & R & C & S \\
\hline
\hline
LSTM \cite{neuraltalk}
& .715 & .545 & .407 & .304 & .248 & .526 & .940 & .178 
& .715 & .544 & .409 & .310 & .249 & .528 & .946 & .178  
& .714 & .543 & .410 & .311 & {\bf .250} & .529 & .951 & {\bf .179} \\ 
CNN 
& .712 & .541 & .404 & .303 & .248 & .527 & .937 & .178 
& .709 & .538 & .403 & .303 & .247 & .525 & .929 & .176 
& .706 & .533 & .400 & .302 & .247 & .522 & .925 & .175 \\ 
CNN+Attn 
& .718 & .549 & .411 & .306 & .248 & .528 & .942 & .177  
& {\bf.722} & {\bf.553} & {\bf.418} & {\bf.316} & {\bf.250} & {\bf.531} & {\bf.952} & {\bf.179} 
& .718 & .550 & .415 & .314 & .249 & .528 & .951 & .179 \\ 
\hline
\end{tabular}}
\end{center}
\vspace{-0.5cm}
\caption{Comparison of different methods (metrics same as Table \ref{tab:metrics}) 
with beam search on the output word probabilities. Our results show that with 
beam size$=3$ our CNN outperforms LSTM \cite{neuraltalk} on all metrics. Note,
compared to \tabref{tab:metrics}, the performance improves with beam search.
We use the MS COCO test split for this experiment. For beam search, we pick 
one caption with maximum log probability (sum of log probability of words) 
from the top-$k$ beams and report the above metrics for it. Beam $=1$ is 
same as the test set results reported in Table \ref{tab:metrics}.}
\label{tab:beam_search}
\end{table*}

\noindent
{\bf CNN Module.} The CNN module operates on the combined input and
image embedding vector. It performs three layers of masked convolutions, 
keeping the feature dimensions after convolution to $512$. 
Consistent with~\cite{S2S,Oord2016}, we use gated linear unit (or GLU) activations for our conv layers. 
However, we did not observe a significant change in performance 
when using the standard ReLU activation. We experiment with weight 
normalization, residual connections and dropout in these layers 
and show that they help improve performance (\tabref{tab:metrics}). 
Our masked convolutions have a receptive field of 5 words in the 
past. We set $N$ (steps or max-sentence length) to $15$ for both CNN/RNN. 
The output of the CNN module after three layers is a $512$-dimensional 
vector for each word. 

\noindent
{\bf Classification Layer.} We use a linear layer to encode the
 $512$-dimensional vectors obtained from the CNN module into a $256$-dimensional representation  
per word. Then, we upsample this vector to a $|\cY|$-dimensional activation via a fully connected layer, and pass it 
through a softmax  to obtain the output word probabilities 
$p_{i,w}(y_{i}|I)$. 

\noindent
{\bf Training.} We use a cross-entropy loss on the probabilities
$p_{i,w}(y_{i}|y_{<i},I)$ to train the CNN module and the embedding layers. Consistent
with~\cite{neuraltalk},  we start to fine-tune VGG16 along with our network
after 8 training epochs. We optimize with RMSProp using an initial learning
rate of $5e^{-5}$ and decay the learning rate by a factor of $.1$ after 
every 15 epochs. All methods were trained for 30 epochs and we evaluate the 
metrics (given in \secref{sec:results_metrics}) on the validation set to 
pick the best model for each method. 

\subsection{Attention}
\label{sec:attn}

In addition to the aforementioned CNN architecture, we also experiment 
with an attention mechanism, since attention benefited~\cite{S2S,VaswaniAttn}. 
We form an attended image vector of dimension $512$ and add it to the 
word embedding at every layer (shown with red, green and blue arrows in
\figref{fig:Detailed}). We compute separate attention parameters and a 
separate attended vector for every word. To obtain this attended vector 
we predict $7 \times 7$ attention parameters, over the VGG16 
max-pooled conv-5 features of dimensions $7 \times 7 \times 512$~\cite{Simonyan14c}.  
We use attention on all three masked convolution layers in our CNN 
module. We continue to use the fc7 image embedding discussed above.

To discuss attention more formally, let $d_{j}$ denote the embedding 
of word $j$ in the conv module (\ie, its activations after GLU shown
in  \figref{fig:Detailed}), let $W$ refer to a linear layer 
applied to $d_j$, let $c_{i}$ denote a $512$-dimensional spatial conv-5 
feature at location $i$ (in $7 \times 7$ feature map) and let 
$a_{ij}$ indicate the attention parameters. With this notation at 
hand, the attention parameter $a_{ij}$ is computed via
\be
a_{ij} = \frac{\exp(W(d_j)^{T}c_i)}{\sum\limits_{i}\exp(W(d_j)^{T}c_i)}, 
\label{eq:attn}
\ee
and the attended image vector for word $j$ is obtained from  
$\sum\limits_{i}a_{ij}c_{i}$. 

Note that~\cite{show_attend_tell} uses the LSTM hidden state to compute the 
attention parameters. Instead, we compute attention parameters using the 
conv-layer activations. This form of attention mechanism was first proposed 
in~\cite{Bahdanau2014NeuralMT}. 

\begin{table*}[!t]
\begin{center}
\setlength{\tabcolsep}{3pt}
\begin{tabular}{|l|ccccccc|ccccccc|}
\hline
\multirow{2}{*}{} & \multicolumn{7}{c|}{c5 (Beam = 1)} & \multicolumn{7}{c|}{c40 (Beam = 1)}\\  
& B1 & B2 & B3 & B4 & M & R & C   
& B1 & B2 & B3 & B4 & M & R & C \\  
\hline
\hline
LSTM 
& .704 & .528 & .384 & .278 & {\bf .241} & {\bf .517} & {\bf .876} 
& .880 & .778 & .656 & .537 & {\bf .321} & .655 & {\bf .898} \\  
CNN+Attn
& {\bf .708} & {\bf .534} & {\bf .389} & {\bf .280} & {\bf .241} & {\bf .517} & .872    
& {\bf .883} & {\bf .786} & {\bf .667} & {\bf .545} & {\bf .321} & {\bf .657} & .893  \\  
\hline
\hline
\multirow{2}{*}{} & \multicolumn{7}{c|}{c5 (Beam = 3)} & \multicolumn{7}{c|}{c40 (Beam = 3)}\\  
& B1 & B2 & B3 & B4 & M & R & C   
& B1 & B2 & B3 & B4 & M & R & C \\  
\hline
\hline
LSTM 
& .710& .537 & .399 & .299 & {\bf .246} & .523 & .904   
& .889& .794 & .681 & .570 & {\bf.334} & .671 & .912 \\  
CNN+Attn
& {\bf .715} & {\bf .545} & {\bf .408} & {\bf .304} & {\bf .246} & {\bf .525} & {\bf .910}   
& {\bf.896} & {\bf .805} & {\bf .694} & {\bf .582} & .333 & {\bf .673} & {\bf .914} \\  
\hline
\end{tabular}
\end{center}
\vspace{-0.5cm}
\caption{Above, we show that CNN outperforms LSTM on BLEU metrics and 
gives comparable scores to LSTM on other metrics for test split on
MSCOCO evaluation server. Here, we submitted our results to the MSCOCO 
evaluation server to evaluate on the hidden test split. Note, this test 
split of $40, 775$ images is different from the $5000$ images test split 
used in Tables \ref{tab:metrics} and \ref{tab:beam_search}. We compare our CNN+Attn method to the LSTM baseline 
on standard evaluation metrics: BLEU-1 (B1), BLEU-2 (B2), BLEU-3 (B3), 
BLEU-4 (B4), METEOR (M), ROUGE (R) and CIDEr (C). The $c5$ scores above 
are computed with $5$ reference captions per test image and $c40$ scores
are computed with $40$ reference captions. We show comparison results for beam 
size 1 and beam size 3 for both the methods.}
\label{tab:cocoeval}
\end{table*}

\section{Results and Analysis}
\label{sec:results}

\begin{table*}[t]
\resizebox{\textwidth}{!}{
\begin{tabular}{lllll}
  \centering
   {\includegraphics[width=.2\textwidth,height=.15\textwidth]{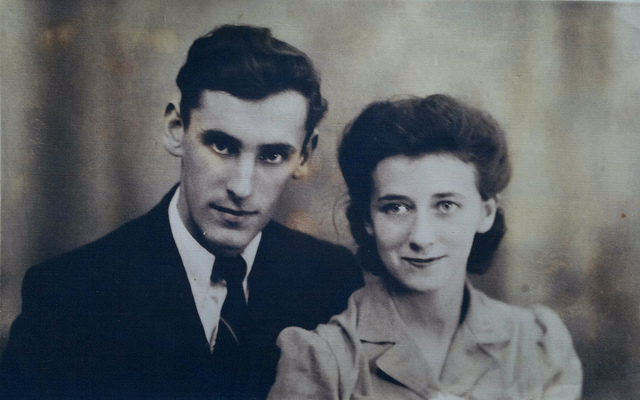}} &
   {\includegraphics[width=.2\textwidth,height=.15\textwidth]{fig/qual_caption/img_2.png}} &
   {\includegraphics[width=.2\textwidth,height=.15\textwidth]{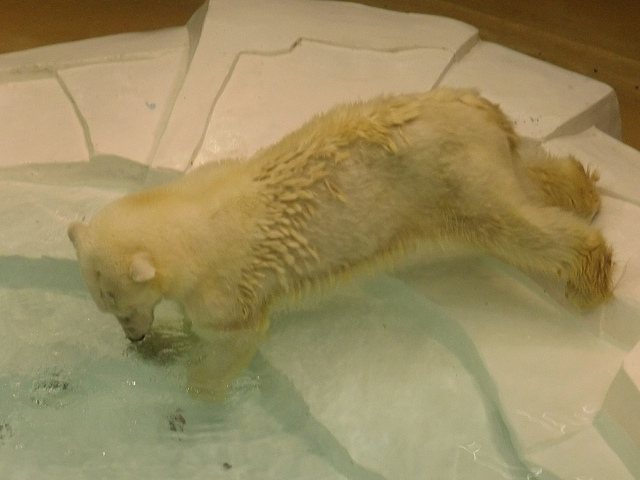}} &
   {\includegraphics[width=.2\textwidth,height=.15\textwidth]{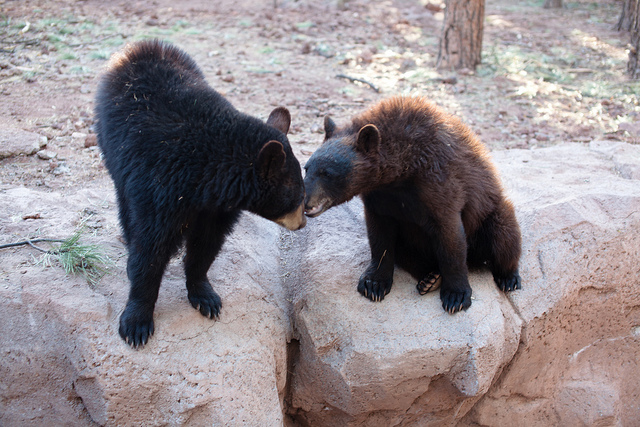}} &
   {\includegraphics[width=.2\textwidth,height=.15\textwidth]{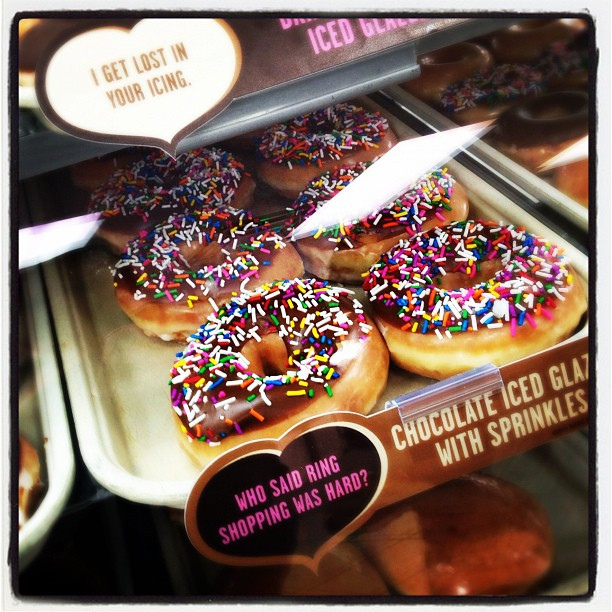}} \\
    \specialcell{
	    {\bf LSTM:} a man and a woman \\ in a suit and tie \\
	    {\bf CNN:} a black and white photo \\ of a man and woman in a suit \\	
	    {\bf GT:} A man sitting next to a \\ woman while wearing a suit.} &
    \specialcell{
	    {\bf LSTM:} a man is walking down \\ a path in the woods \\
	    {\bf CNN:}  a truck is parked in \\ the dirt near a tree\\	
	    {\bf GT:} a truck is parked at a \\ campground with snow on it	} &
    \specialcell{
	    {\bf LSTM:} a cat is laying \\ down on a bed	 \\
	    {\bf CNN:}  a polar bear is drinking \\ water from a white bowl	\\	
	    {\bf GT:} A white polar bear laying \\ on top of a pool of water} &
    \specialcell{
	    {\bf LSTM:}  a bear is standing \\ on a rock in a zoo	\\	
	    {\bf CNN:}  two bears are walking \\ on a rock in the zoo	\\
	    {\bf GT:} two bears touching \\ noses standing on rocks} &
    \specialcell{
	    {\bf LSTM:}  a box of donuts with \\ a variety of toppings	\\
	    {\bf CNN:}  a box of doughnuts with \\ sprinkles and a sign \\	
	    {\bf GT:}A bunch of doughnuts \\ with sprinkles on them} \\ 
   {\includegraphics[width=.2\textwidth,height=.15\textwidth]{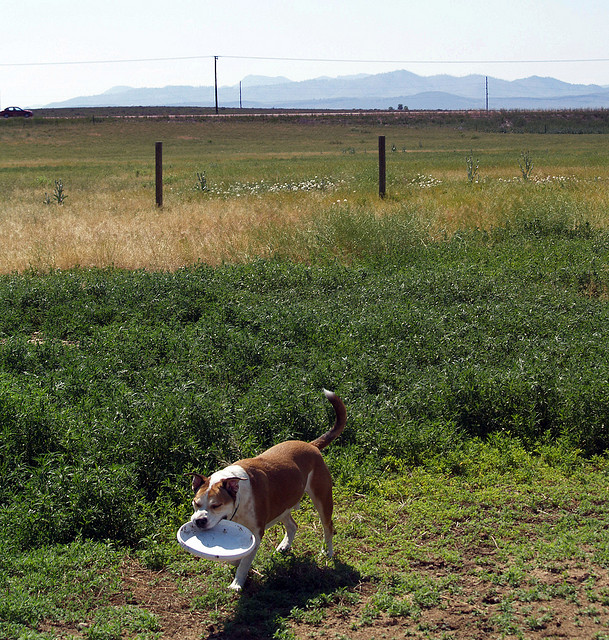}} &
   {\includegraphics[width=.2\textwidth,height=.15\textwidth]{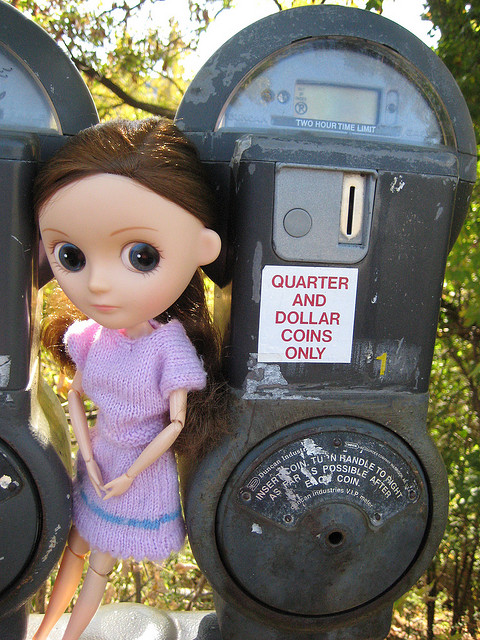}} &
   {\includegraphics[width=.2\textwidth,height=.15\textwidth]{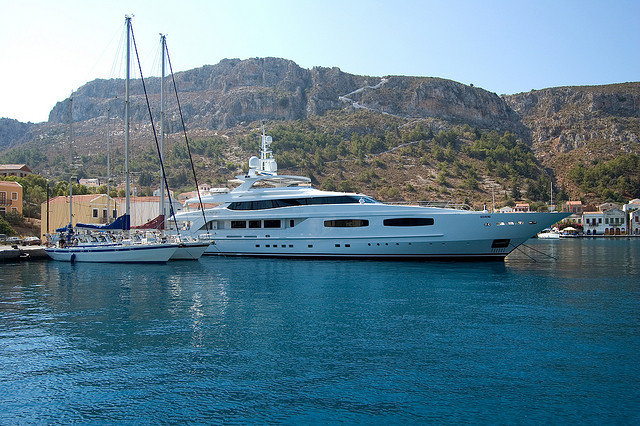}} &
   {\includegraphics[width=.2\textwidth,height=.15\textwidth]{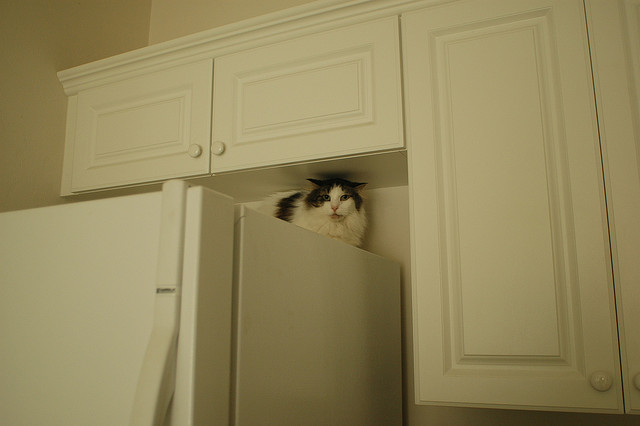}} &
   {\includegraphics[width=.2\textwidth,height=.15\textwidth]{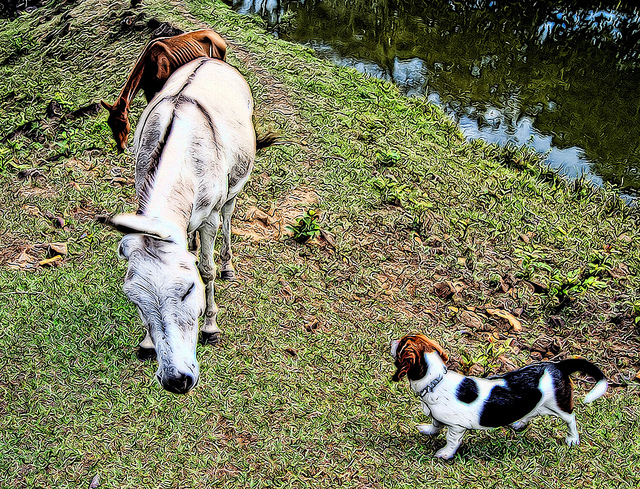}} \\
    \specialcell{
	    {\bf LSTM:} a dog is standing in \\ the grass near a tree\\
	    {\bf CNN:}  a dog is standing in \\ the grass  with a frisbee \\ in its mouth	 \\	
	    {\bf GT:} large dog retrieving \\ the frisbee for his owner} & 
    \specialcell{
	    {\bf LSTM:} a parking meter \\ with a sign on it	\\
	    {\bf CNN:}  a doll is sitting \\ next to a parking meter \\	
	    {\bf GT:} A doll with articulated \\ joints stares from her perch \\ between two parking meters.} & 
    \specialcell{
	    {\bf LSTM:}   a boat is docked in \\ the water near a dock \\	
	    {\bf CNN:} a boat is docked in \\ the water near a mountain	\\
	    {\bf GT:} A white boat floating on \\ a lake under mountains} & 
    \specialcell{
	    {\bf LSTM:} a cat is standing on \\ top of a refrigerator	\\
	    {\bf CNN:}  a cat is standing in a \\ kitchen looking at the camera \\	
	    {\bf GT:}  A cat in a kitchen \\ on top of a refrigerator }& 
    \specialcell{
	    {\bf LSTM:} a dog and a \\ dog in a field	\\
	    {\bf CNN:} two cows are  \\ standing  in a field of grass \\	
	    {\bf GT:} A dog and a horse \\ standing near each other} \\ 

\end{tabular}}
  \captionof{figure}[]{Captions generated by our CNN are compared to the LSTM and 
  ground-truth caption. In the examples above our CNN can describe things like black and white photo, 
  polar bear/white bowl, number of bears, sign in the donut image which LSTM fails
  to do. The last two images (bottom right) show failure cases for CNN. Typically we observe that CNN and LSTM captions are of similar quality. We use our 
  CNN+Attn method (\secref{sec:attn}) and the MSCOCO test split for these results.}
  \label{fig:qual}
\end{table*}

In this section, we demonstrate the following results:  

\begin{itemize}
\setlength{\itemsep}{0pt}
 \item Our convolutional (or CNN) approach performs on par with 
       LSTM (or RNN) based approaches on image captioning metrics 
       (\tabref{tab:metrics}). Our performance improves with 
       beam search (\tabref{tab:beam_search}). 
 \item Adding attention to our CNN gives  improvements on metrics 
       and we outperform the LSTM+Attn baseline~\cite{show_attend_tell}
       (\tabref{tab:metrics}). \figref{fig:attention} shows that with 
       attention we identify salient objects for the given image. 
 \item We analyze the CNN and RNN approaches and show that CNN
       produces (1) more entropy in the output probability distribution,
       (2) gives better word prediction accuracy (\figref{fig:analysis}), and (3) does not suffer
       as much from vanishing gradients (\figref{fig:grad}). 
  \item In \tabref{tab:time}, we show that  a CNN with $1.5 \times$ 
  	more parameters can be trained in comparable time. This is
	because we avoid the sequential processing of RNNs.
\end{itemize}

The details of our experimental setup and these results 
are discussed below. PyTorch implementation of our
convolutional image captioning is available on github.\footnote{\url{https://github.com/aditya12agd5/convcap}}

\begin{table*}[t]
\vspace{0.5cm}
\resizebox{\textwidth}{!}{
\begin{tabular}{lccccccc}
  \centering
   {\includegraphics[width=.2\textwidth]{fig/qual_attention/img_002_018.png}} &
   {\includegraphics[width=.2\textwidth]{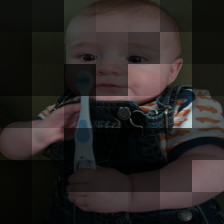}} &
   {\includegraphics[width=.2\textwidth]{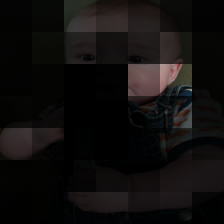}} &
   {\includegraphics[width=.2\textwidth]{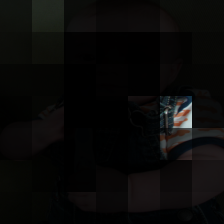}} &
   {\includegraphics[width=.2\textwidth]{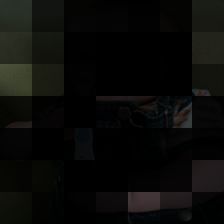}} &
   {\includegraphics[width=.2\textwidth]{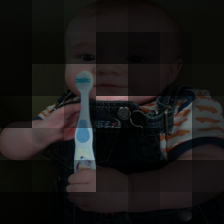}} &
   {\includegraphics[width=.2\textwidth]{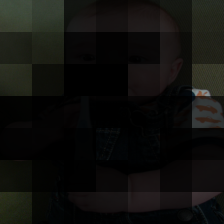}} &
   {\includegraphics[width=.2\textwidth]{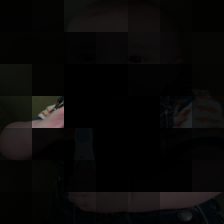}} \\
    \specialcell{{\bf CNN:}  a baby holding a \\ toothbrush in his hand   \\
	    {\bf GT:} A child holds a \\ toothbrush in their hand.} & 
    a & baby & holding & a & toothbrush & in & ... hand \\
   {\includegraphics[width=.2\textwidth]{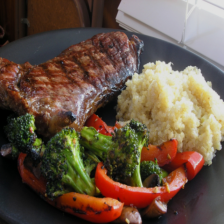}} &
   {\includegraphics[width=.2\textwidth]{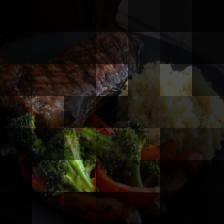}} &
   {\includegraphics[width=.2\textwidth]{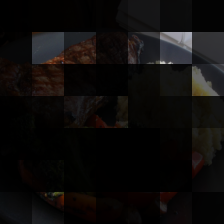}} &
   {\includegraphics[width=.2\textwidth]{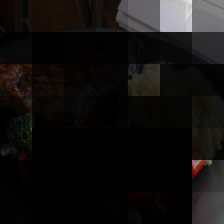}} &
   {\includegraphics[width=.2\textwidth]{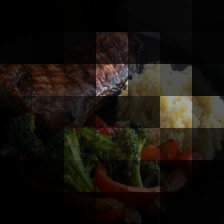}} &
   {\includegraphics[width=.2\textwidth]{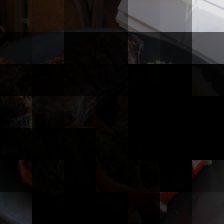}} &
   {\includegraphics[width=.2\textwidth]{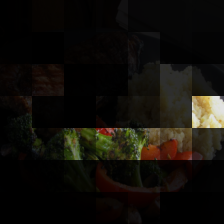}} &
   {\includegraphics[width=.2\textwidth]{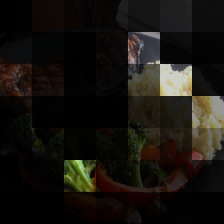}} \\
    \specialcell{{\bf CNN:}   a plate of food with \\ broccoli and rice    \\
	    {\bf GT:} A BBQ steak on a plate \\ next to mashed potatoes \\ and mixed vegetables. } & 
     a & plate & of & food & with & broccoli & ... rice   \\
   {\includegraphics[width=.2\textwidth]{fig/qual_attention/img_004_015.png}} &
   {\includegraphics[width=.2\textwidth]{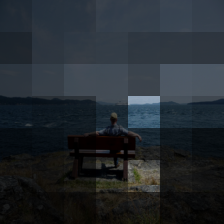}} &
   {\includegraphics[width=.2\textwidth]{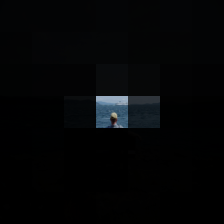}} &
   {\includegraphics[width=.2\textwidth]{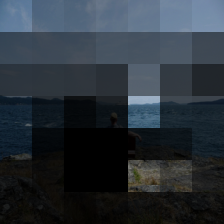}} &
   {\includegraphics[width=.2\textwidth]{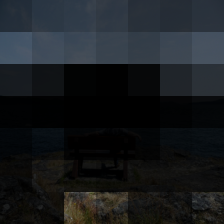}} &
   {\includegraphics[width=.2\textwidth]{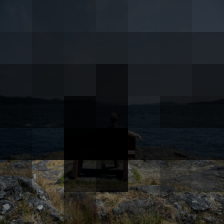}} &
   {\includegraphics[width=.2\textwidth]{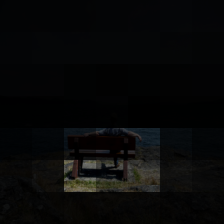}} &
   {\includegraphics[width=.2\textwidth]{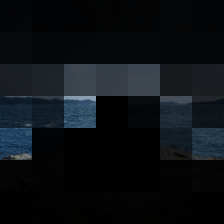}} \\
    \specialcell{{\bf CNN:} a man sitting on a \\ bench overlooking the ocean        \\
	    {\bf GT:} A man sitting on top \\ of a bench near the ocean}  & 
     a & man & sitting & on & a & bench & ... ocean    \\ 
\end{tabular}}
\vspace{-0.3cm}
  \captionof{figure}[]{We overlay the attention parameters on the image. These results
  show that we focus on salient objects such as broccoli, bench, toothbrush in the image 
  when predicting these words. Also, the attention is spread out/uniform when predicting
  words such as a, of and on which are unrelated to image content.}
  \label{fig:attention}
  \vspace{-0.5cm}
\end{table*}

\subsection{Dataset and Baselines}
We conducted experiments on the \textbf{MS COCO} dataset~\cite{MSCOCO}. 
Our train/val/test splits follow~\cite{neuraltalk,show_attend_tell}. 
We use 113287 training images, 5000 images for 
validation, and 5000 for testing. Henceforth, we will refer to our 
approach as CNN, and our approach with the attention (\secref{sec:attn}) 
as CNN+Attn. We use the following naming convention for our 
baselines: \cite{neuraltalk} is denoted by LSTM and 
\cite{show_attend_tell} is referred to as LSTM+Attn.

\subsection{Comparison on Image Captioning Metrics}
\label{sec:results_metrics}

We consider multiple conventional evaluation metrics,  
BLEU-1, BLEU-2, BLEU-3, BLEU-4 \cite{BLEU}, METEOR \cite{METEOR}, 
ROUGE \cite{ROUGE}, CIDEr \cite{Rama} and SPICE \cite{SPICE}. 
See \tabref{tab:metrics} for the performance on all 
these metrics for our val/test splits. Note that we obtain
comparable CIDEr scores and better SPICE scores than 
LSTM on test set with our CNN+Attn method. Our BLEU, METEOR, 
ROUGE scores are less than the LSTM ones, but the margin is
very small. Our CNN+Attn method outperforms the LSTM+Attn 
baseline on the test set for all metrics reported in~\cite{show_attend_tell}. For \tabref{tab:metrics}, we 
form the caption by choosing the word with maximum 
probability at each step. The metrics are reported for 
this one caption formed by choosing maximum probability 
word at every step.

Instead of sampling the maximum probability words, we 
also perform  beam search with different beam sizes. 
We perform beam search for both LSTM and our CNN methods.
With beam search, we pick the maximum probability 
caption (sum of log word probability in the beam). The 
results reported in  \tabref{tab:beam_search} demonstrate 
that with beam size of $3$ we achieve better BLEU, ROUGE, 
CIDEr scores than LSTM and equal METEOR and SPICE 
scores. 

In Table~\ref{tab:cocoeval}, we show the results obtained
on the MSCOCO evaluation server. These results are computed
over a test set of $40,775$ images for which ground-truth
is not publicly available. We demonstrate that our method does 
better on all BLEU metrics, especially with beam size 3, we 
perform better than the LSTM based method.

\subsection{Qualitative Comparison}
\label{sec:qual}

See \figref{fig:qual} for a qualitative comparison of 
captions generated by CNN and LSTM. In 
\figref{fig:attention}, we overlay the attention 
parameters on the image for each word prediction. 
Note that our attention parameters are $7 \times 7$
as described in \secref{sec:attn} and therefore
the image is divided in a $7 \times 7$ grid. These 
results show that our attention 
focuses on salient objects such as man, broccoli, ocean, 
bench, \etc, when predicting these respective words.
Our results also show that the attention is uniform 
when predicting words such as a, of, on, \etc, 
which are unrelated to the image content. 

\subsection{Analysis of CNN and RNN}
In \tabref{tab:time} we report the number of trainable parameters and the training time per epoch.  CNNs with $1.5$ times more parameters can be trained in comparable time. 

\begin{figure*}[t]
  \centering
  \subfigure[CNN gives higher cross-entropy loss on train/val set of MSCOCO 
  compared to LSTM. But, as we show in (c), our CNN obtains better $\%$ word 
  accuracy than LSTM. Therefore, it still assigns max. probability to correct 
  word. The CNN loss is high because our output probability distributions have more
  entropy than LSTM.]{\includegraphics[width=.3\textwidth]{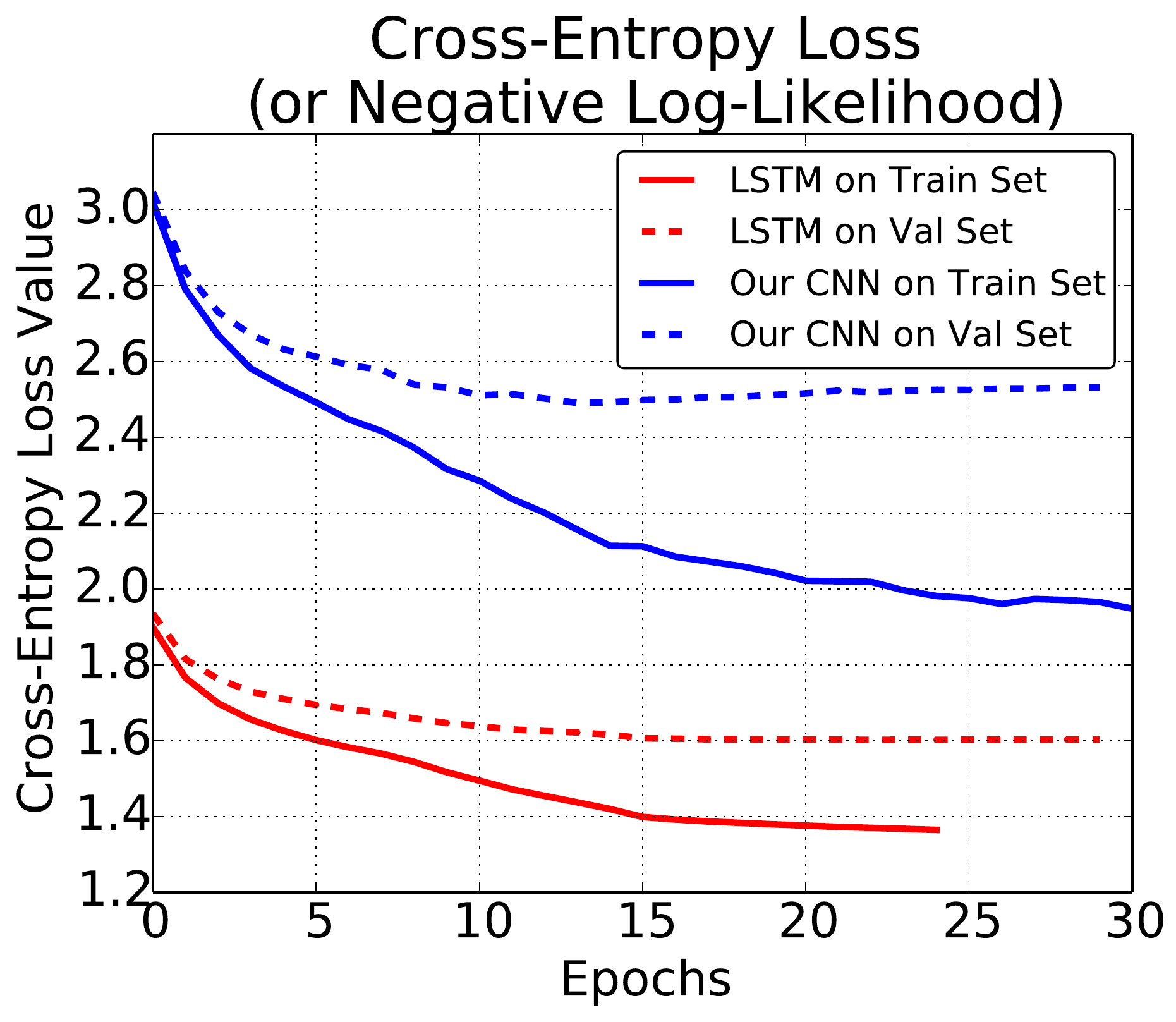}} \quad 
  \subfigure[The entropy of the softmax layer (or posterior probability distribution) of our CNN is higher than 
  	the LSTM. For ambiguous problems such as image captioning, it is desirable to have a less peaky (multi-modal) 
	posterior (like ours) capable of producing multiple captions, rather than a peaky one (like LSTM).]{\includegraphics[width=.3\textwidth]{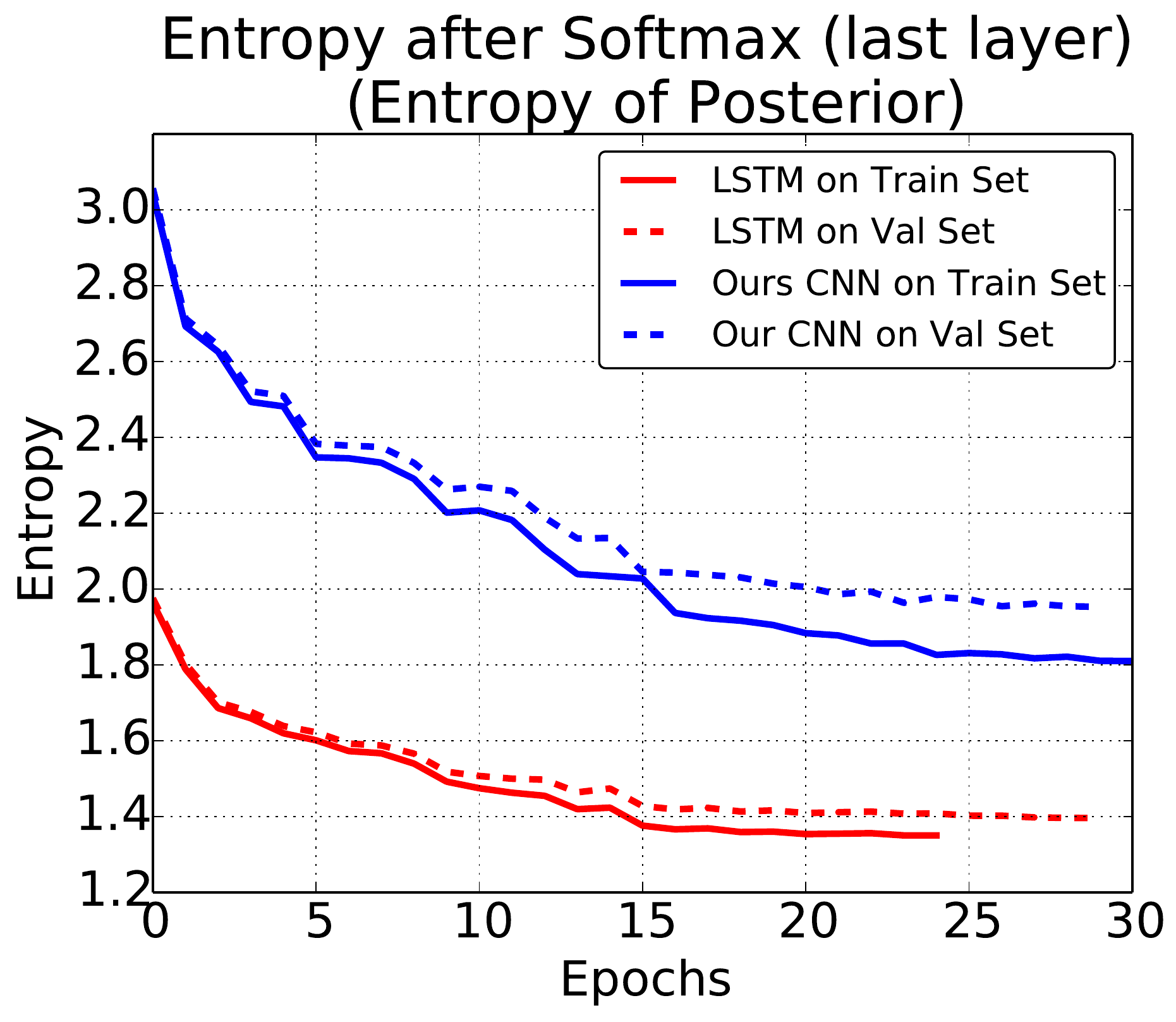}} \quad 
  \subfigure[Even though the CNN training loss is higher than LSTM, its word prediction accuracy
  	is better than LSTM on train set. On val set, the difference in accuracy between LSTM and CNN is small
	(only $\sim 1\%$).]{\includegraphics[width=.3\textwidth]{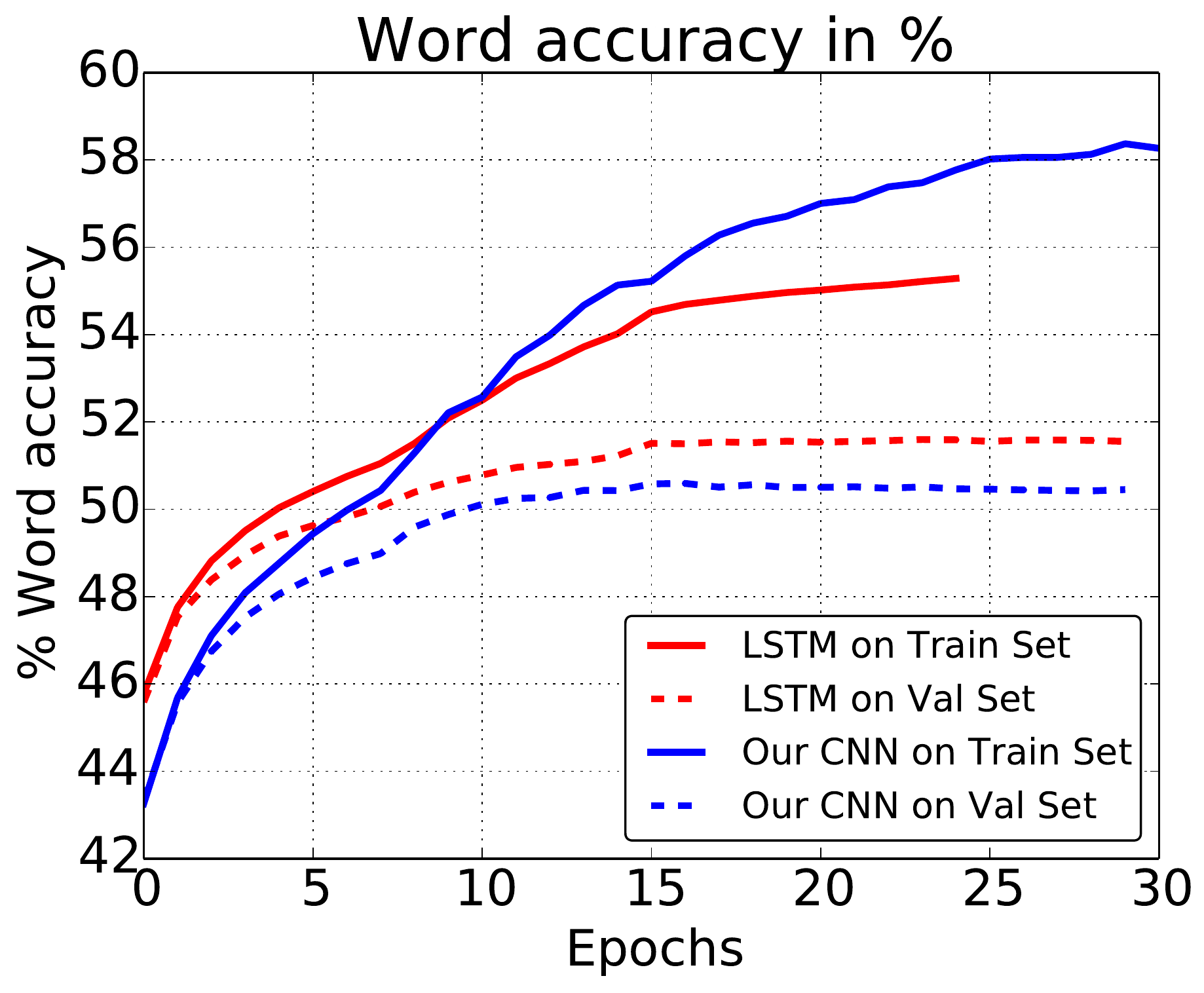}} \quad
	\vspace{-0.4cm}
  \caption{In the figures above we plot (a) Cross-entropy loss, (b) Entropy of the softmax layer,
  (c) Word accuracy on train/val set. Blue line denotes our method and red denotes the LSTM based method 
  \cite{neuraltalk}. In (a), (b) and (c) solid/dotted lines denote train/val set of MSCOCO 
  respectively. For these plots, at the end of every epoch we randomly sample $10$k images from train set 
  and use the entire val set.}
  \label{fig:analysis}
  \vspace{-0.5cm}
\end{figure*}

\tabref{tab:metrics}, \ref{tab:beam_search} and \ref{tab:cocoeval} show
that we obtain comparable performance from both CNN
and RNN/LSTM-based methods. Encouraged by this result, we analyze the characteristics of 
these two methods. For fair comparison, we use our CNN without attention, since
the RNN method does not use spatial image features. First, we compare the negative log-
likelihoods (or cross-entropy loss) on a subset of train 
and the entire val set (see \figref{fig:analysis} (a)). We 
find that the loss is higher for CNN than RNN, this is because CNNs are being penalized for producing less-peaky word probability distributions. To evaluate this further, we plot the entropy of the output probability distribution 
(\figref{fig:analysis} (b)) and the classification accuracy,
\ie, the number of times the maximum probability word is the ground truth 
(\figref{fig:analysis} (c)). These plots show that RNNs
are good at producing low entropy and therefore peaky word
probability distributions at the output, while CNNs produce 
less peaky distributions (and high entropy). Less peaky
distributions are not necessarily bad, particularly for
a problem like image captioning, where multiple 
word predictions are possible. Despite, less peaky 
distributions, \figref{fig:analysis}(c) shows that the
maximum probability word is correct more often on the 
train set and it is within approx.~$1\%$ accuracy on the val set. 
Note, cross-entropy loss is a proxy for the classification
accuracy and we show that CNNs have higher cross entropy
loss, but their classification accuracy is good. Less 
peaky posterior distributions provided by a CNN may be 
indicative of CNNs being more capable of predicting diverse 
captions. 

\begin{figure}[!t]
  \includegraphics[width=.45\textwidth]{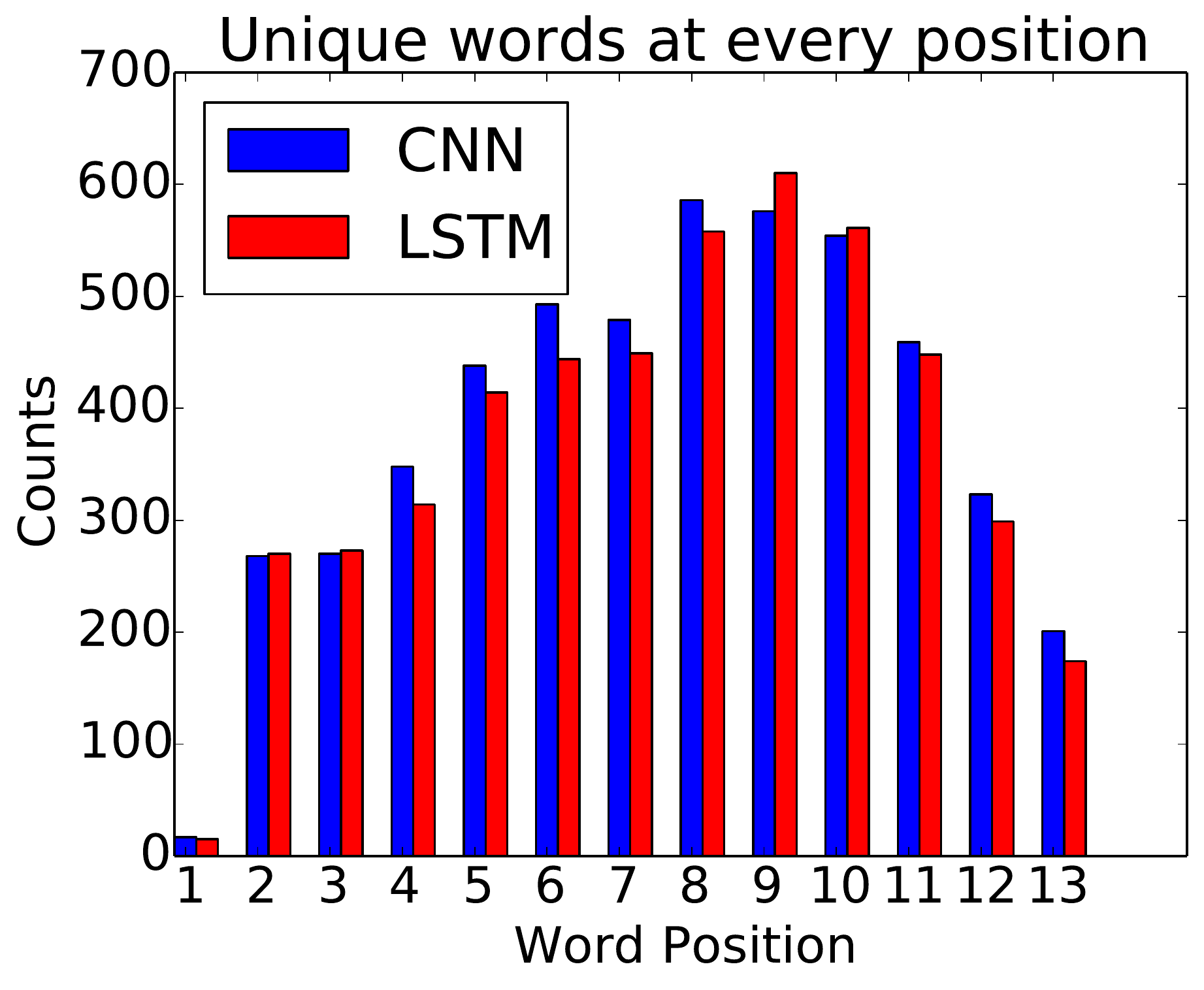}
  \vspace{-0.5cm}
  \caption{We perform beam search of beam size 10 with our best performing LSTM
   and CNN models. We use the top 10 beams to plot the unique words predicted for 
   every word position. CNN produces higher unique words at more word positions
   than LSTM.} 
  \label{fig:words}
\end{figure}

In Figure \ref{fig:words}, we plot the unique words predicted at every
word position or time-step. The plot is for word positions $1$ to $13$.
This plot shows that for the CNN we have higher unique words for more word positions
than LSTM. This supports our analysis that CNNs have
less peaky (or one-hot) posteriors and therefore can produce more diversity. For
these diversity experiments, we perform a beam search with beam size 10 and use
all the top 10 beams.

\begin{figure}[!t]
  \includegraphics[width=.45\textwidth]{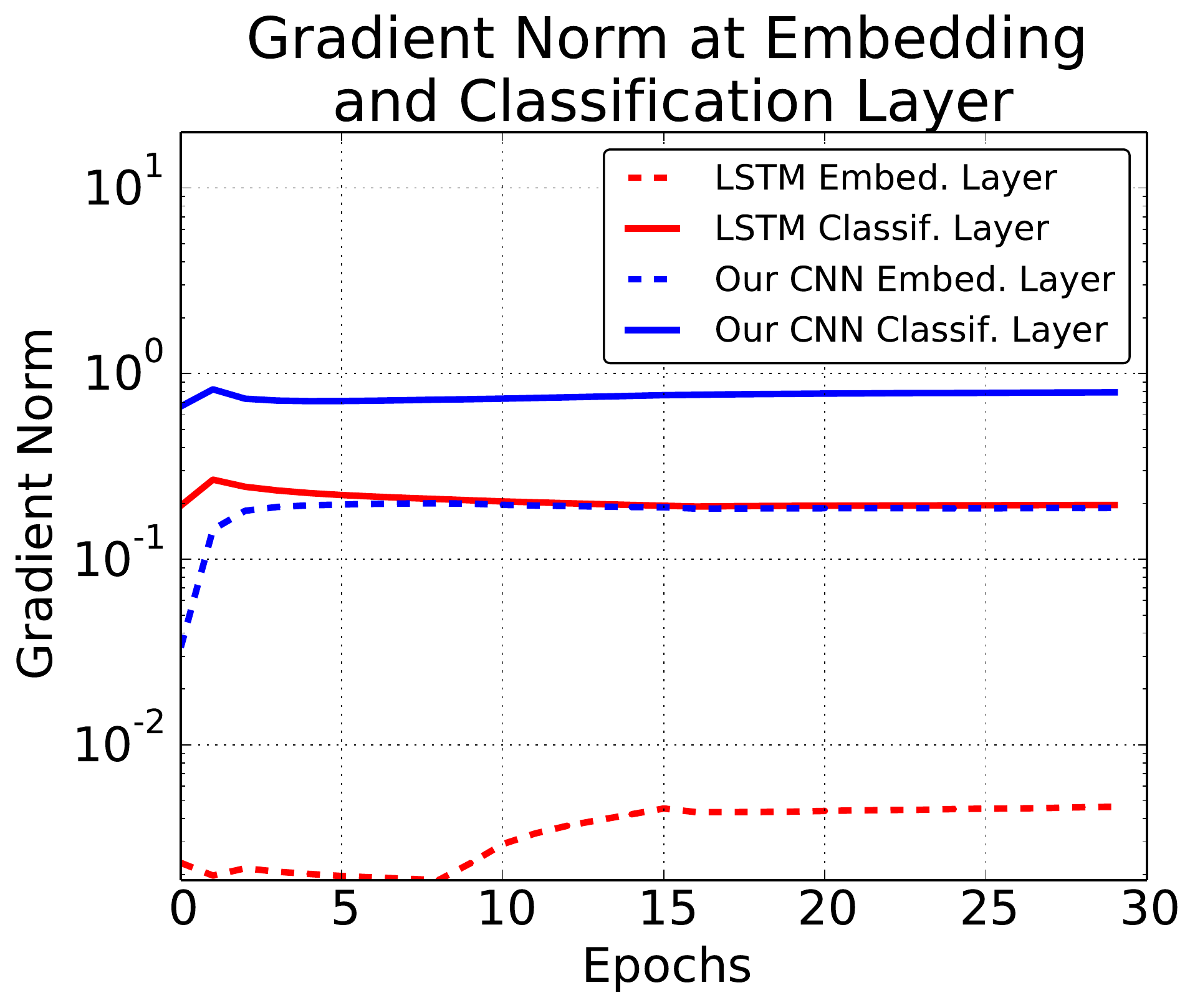}
  \vspace{-0.5cm}
  \caption{Here, we plot the gradient norm at the input embedding (first) and output embedding 
	  (last/classification) layer. The gradient to the first layer of LSTM decays by a 
	   factor $\sim 100$ in contrast to our CNN, where it decays by a factor of $\sim 10$. 
	   There is prior evidence in literature that unlike CNNs, RNN/LSTMs suffer from 
	   vanishing gradients \cite{PascanuDiffRNN,ICML2011Sutskever_524}. Here, we use solid 
	   line for input embedding layer and dotted line for the classification layer.}
  \label{fig:grad}
  \vspace{-0.5cm}
\end{figure}

Since RNNs/LSTMs are known to suffer from vanishing gradient
problems, in \figref{fig:grad}, we plot the gradient 
norm at the output embedding/classification layer and the 
gradient norm at the input embedding layer. The values are 
averaged over 1 training epoch. These plots show that
the gradients in RNN/LSTM diminishes more than the ones in CNNs.
Hence RNN/LSTM nets are more likely to suffer from vanishing 
gradients, which stalls learning. If learning is stalled, 
for larger datasets than the ones we currently use for 
image captioning, the performance of RNN and CNN may
 differ significantly.

\begin{table}[!t]
\begin{center}
\setlength{\tabcolsep}{3pt}
\begin{tabular}{|c|c|c|}
\hline
\bf{Method} & {\bf \# Parameters} & \bf{Train time per epoch} \\ 
\hline
\hline
LSTM \cite{neuraltalk} & ~13M &  1529s \\
Our CNN &  ~19M & 1585s \\
Our CNN+Attn & ~20M & 1620s \\
\hline
\end{tabular}
\end{center}
\vspace{-0.5cm}
\caption{Comparison of train time (in seconds) for LSTM and CNN.
We train a CNN faster per parameter than the LSTM. This is 
because CNN is not sequential like the LSTM. We use PyTorch 
implementation of \cite{neuraltalk} and our CNN-based method for 
fair comparison. The timings are obtained on Nvidia Titan X GPU.}
\label{tab:time}
\vspace{-0.5cm}
\end{table}

 \section{Related Work}
 \label{sec:relwork}

Describing the content of an observed image is related to a large variety of
tasks. Object detection~\cite{YOLO,fasterRCNN} and semantic
segmentation~\cite{Zoomout,semantic_segmentation} can be used to obtain a list of
objects. Detection of co-occurrence patterns and relationships between
objects can help to form sentences. Generating sentences by taking advantage of 
surrogate tasks is then a multi-step approach which is beneficial for interpretability 
but lacks a joint objective that can be trained end-to-end. 

Early techniques formulate image captioning as a retrieval problem and find the
best fitting description from a pool of possible 
captions~\cite{retrieval1,retrieval5,retrieval4,retrieval2}. Those
techniques are built upon the idea that the fitness between available textual
descriptions and images can be learned. While this permits end-to-end training,
matching image descriptors to a sufficiently large pool of captions
is computationally expensive. In addition, constructing a database
of captions that is sufficient for describing a reasonably large
fraction of images seems prohibitive.

To address this issue, recurrent neural nets (RNNs) or probabilistic models like
Markov chains, which decompose the space of a caption into a product space of
individual words are compelling. The success of RNNs for image captioning is based 
on a key component, \ie, the Long-Short-Term-Memory (LSTM)~\cite{HochreiterLSTM} 
or recent alternatives like the gated recurrent unit (GRU)~\cite{ChoGRU}. 
These components capture long-term dependencies by adding a memory cell, and 
they address the vanishing or exploding gradient issue of classical RNNs to some degree. 

Based on this success,~\cite{Mao2014DeepCW} train a vision (or image) CNN and a 
language RNN that shares a joint embedding layer.~\cite{show_tell} jointly train a 
vision (or image) CNN with a language RNN to generate sentences,~\cite{show_attend_tell} 
extends~\cite{show_tell} with additional attention parameters and learns to identify 
salient objects for caption generation.~\cite{neuraltalk} use a bi-directional RNN
along with a structured loss function in a shared vision-language space. These 
recurrent neural nets have found widespread use for captioning because they have 
been shown to produce remarkably fitting descriptions.

Despite the fact that the above RNNs based on LSTM  or GRU units deliver remarkable
results, \eg, for image captioning, their training procedure is all but trivial. 
For instance, while the forward pass during training can be parallelized across
samples, it is inherently sequential in time, limiting the parallelization
abilities. To address this issue,~\cite{Oord2016} proposed a PixelCNN architecture 
for conditional image generation that approximates an RNN for the same task.
\cite{S2S} and~\cite{VaswaniAttn} demonstrate that convolutional architectures 
with attention achieve state-of-the-art performance on machine translation 
tasks. In spirit similar is our approach for image captioning, which is 
convolutional but addresses a different task. 


\section{Conclusion}
We discussed a convolutional approach for image captioning and showed that it performs on par with existing LSTM techniques. We also analyzed the differences between RNN based learning and our method, and found gradients of lower magnitude as well as overly confident predictions to be existing LSTM network concerns.\\

\section{Acknowledgments}
\vspace{-0.3cm}
\noindent We thank Arun Mallya for providing us with an efficient PyTorch 
implementation of \cite{neuraltalk}, we thank Tanmay Gangwani for 
the beam search code used for Figure \ref{fig:words} and we thank 
David Forsyth for insightful discussions and his comments. This material is based upon work supported in part by the National Science Foundation under Grant No.~1718221.


{\small
\bibliographystyle{ieee}
\bibliography{egbib}
}

 \end{document}